\RequirePackage{fix-cm}
\documentclass[smallextended]{svjour3}  
\smartqed
\usepackage[T1]{fontenc}
\usepackage{graphicx}
\usepackage{subfigure}
\usepackage{cite}
\usepackage{tipa}
\usepackage{url}
\usepackage{float}
\usepackage{amsmath,amssymb,amsfonts}
\usepackage{algorithmic}
\usepackage{graphicx}
\usepackage{textcomp}
\usepackage{adjustbox}
\usepackage[graphicx]{realboxes}
\usepackage{rotating}
\usepackage[misc]{ifsym}
\usepackage{setspace}
\usepackage{algorithm}
\usepackage{algorithmic}

\begin{document}

\title{Dataset Complexity Assessment Based on Cumulative Maximum Scaled Area Under Laplacian Spectrum}

\titlerunning{Dataset Complexity Assessment Based on cmsAULS}

\author{Guang Li $^{1}$ \and Ren Togo $^{2}$ \and Takahiro Ogawa $^{3}$ \and Miki Haseyama $^{3}$
}

\institute{Guang Li $\textrm{\Letter}$ \\
           guang@lmd.ist.hokudai.ac.jp \\ \\
           Ren Togo \\
           togo@lmd.ist.hokudai.ac.jp \\ \\
           Takahiro Ogawa \\
           ogawa@lmd.ist.hokudai.ac.jp \\ \\
           Miki Haseyama \\
           miki@ist.hokudai.ac.jp \\ \\
           $^{1}$ 
              Graduate School of Information Science and Technology, Hokkaido University, Sapporo, Japan \\ \\
           $^{2}$
              Education and Research Center for Mathematical and Data Science, Hokkaido University, Sapporo, Japan \\ \\
           $^{3}$
              Faculty of Information Science and Technology, Hokkaido University, Sapporo, Japan 
}

\date{Received: 30 May 2021 / Revised: 12 October 2021 / Accepted: 28 March 2022}

\maketitle
\begin{abstract}
Dataset complexity assessment aims to predict classification performance on a dataset with complexity calculation before training a classifier, which can also be used for classifier selection and dataset reduction.
The training process of deep convolutional neural networks (DCNNs) is iterative and time-consuming because of hyperparameter uncertainty and the domain shift introduced by different datasets.  
Hence, it is meaningful to predict classification performance by assessing the complexity of datasets effectively before training DCNN models.
This paper proposes a novel method called cumulative maximum scaled Area Under Laplacian Spectrum (cmsAULS), which can achieve state-of-the-art complexity assessment performance on six datasets.
\end{abstract}

\section{Introduction}
With the development of deep convolutional neural networks (DCNNs)~\cite{krizhevsky2012imagenet}, the classification performance of DCNN-based methods has significantly improved. 
However, training DCNN models requires a massive amount of computation time~\cite{liu2019rethinking}, and we cannot confirm test classification performance before the training process because of the uncertainty of DCNN models~\cite{gal2016dropout}. 
Because of the high correlation between the classification performance of DCNN models and the complexity of datasets, some complexity assessment methods have been proposed to solve the aforementioned problems~\cite{lorena2019complex}.
By effectively evaluating a dataset's complexity in advance, we can estimate the classification performance of DCNN models trained on the dataset, saving a substantial amount of time~\cite{li2020complexity}. 
Furthermore, complexity assessment methods can be used in certain applications ($e.g.$, classifier selection~\cite{brun2018framework} and dataset reduction~\cite{leyva2014set}).
\par
Dataset complexity assessment methods aim to evaluate the entanglement degree of dataset classes.
The most well-known complexity assessment method proposed in~\cite{ho2002complexity} has 12 different descriptors, including feature overlap methods, linearity methods, neighborhood methods, and dimension methods~\cite{pascual2020revisiting}.
For example, some descriptors assume that datasets with small overlapping classes are easier to classify than those with large overlaps.   
Since these descriptors assume that classes are linearly separable in their original feature space, they are less suited for analyzing large and complex image datasets~\cite{brun2018framework}.
While some complexity assessment methods designed for two-class classification problems have been validated on some high-dimensional biomedical datasets~\cite{baumgartner2006data, anwar2014measurement}, these methods cannot deal with a multiclass classification problem.
Moreover, some methods require high-dimensional matrix analysis, and are hence memory intensive and time-consuming~\cite{duin2006object, hoiem2012diagnosing}.
\par
Recently, the cumulative spectral gradient method (CSG)~\cite{branchaud2019spectral} has achieved state-of-the-art performance in dataset complexity assessment.
The method focuses on the overlap in feature distribution between image dataset classes and specifically calculates dataset complexity based on the eigenvalues of a Laplacian matrix derived from the similarity matrix between the classes.
A large Laplacian spectrum can denote a large class overlap and can be used as a complexity assessment method~\cite{von2007tutorial}. 
Although CSG pays attention to the gradient between adjacent eigenvalues ($i.e.$, eigengap) of the Laplacian matrix, it does not fully account for the effect of the Area Under Laplacian Spectrum (AULS), which can also influence the spectrum's size.
\par
In this paper, we propose a novel method to improve performance in evaluating image dataset complexity.
From spectral clustering theory~\cite{von2007tutorial}, the Laplacian spectrum size can denote similarities between dataset classes and can therefore be used to assess dataset complexity~\cite{mohar1997some}.
Moreover, two elements can affect Laplacian spectrum size, the AULS and the gradient between adjacent eigenvalues.
The previous methods only focus on one of the AULS and the gradient between adjacent eigenvalues.
However, our proposed dataset complexity assessment method called cumulative maximum scaled Area Under Laplacian Spectrum (cmsAULS) focuses on both the AULS and the gradient between adjacent eigenvalues, achieving better assessment performance than that of previous methods.
We performed experiments on six datasets and achieved state-of-the-art performance in dataset complexity assessment.
\par
Our contributions are summarized as follows:
\begin{itemize}
    \item We propose a new dataset complexity assessment method (cmsAULS) that focuses on both the gradient between the adjacent eigenvalues of a Laplacian spectrum and the area under it.
    \item We confirm that our method outperforms other state-of-the-art dataset complexity assessment methods on six datasets.
\end{itemize}
\par
The rest of this paper is organized as follows.
Related works and our cmsAULS are presented in sections 2 and 3, respectively.
Experiments and Discussion to verify the effectiveness of the proposed method are shown in sections 4 and 5, respectively. 
The conclusion is given in section 6.
\section{Related works}
\begin{table}[t]
    \centering
    \caption{Characteristics of the 12 descriptors for complexity assessment~\cite{lorena2019complex}.}
    \label{tab1}
    \begin{tabular}{c|c|c}
    \hline
    Name & Description & Asymptotic \\\hline\hline
    F1 & Maximum Fisher's discriminant & $O(N\cdot d)$ \\
    F2 & Volume of overlapping region & $O(N\cdot d \cdot n)$ \\
    F3 & Maximum individual feature efficiency & $O(N\cdot d \cdot n)$ \\\hline
    N1 & Fraction of borderline points & $O(N\cdot d^{2})$ \\
    N2 & Ratio of intra/extra class NN distance & $O(N\cdot d^{2})$ \\
    N3 & Error rate of NN classifier & $O(N\cdot d^{2})$ \\
    N4 & None linearity of NN classifier & $O(N\cdot d^{2} + N\cdot l\cdot d)$ \\\hline
    L1 & Sum of the error distance by linear programming & $O(d^{2})$ \\
    L2 & Error rate of linear classifier & $O(d^{2})$ \\
    L3 & Non linearity of linear classifier & $O(d^{2} + N\cdot l\cdot n)$ \\\hline
    T1 & Fraction of hyperspheres covering data & $O(N \cdot d^{2})$ \\
    T2 & Average number of features per dimension & $O(N + d)$ \\
    \hline
    \end{tabular}
\end{table}
For the dataset complexity assessment task, the seminal methods include 12 descriptors (F1, F2, F3, N1, N2, N3, N4, L1, L2, L3, T1, and T2) proposed in this paper~\cite{ho2002complexity}.
F1 is the maximum Fisher’s discriminant ratio.
F2 calculates the interclass overlap of feature distributions.
F3 measures each feature's efficiency in separating the classes and finds the maximum value.
F1, F2, and F3 are feature-based methods that characterize how informative the available features are to separate the classes.
N1, N2, N3, and N4 are neighborhood methods that describe the presence and density of the same or different classes in local neighborhoods.
Meanwhile, L1, L2, and L3 are linearity methods that quantify whether the classes can be separated linearly.
T1 is regarded as a topological method that measures the total number of hyperspheres one can fit into the feature space of a class, while T2 divides the number of examples in the dataset by their dimension.
Characteristics of the 12 descriptors for complexity assessment are summarized in Table~\ref{tab1}.
We also present the asymptotic time complexity of these descriptors, where $N$ stands for the number of points in a dataset, $d$ corresponds to its number of features, $n$ is the number of classes, and $l$ is the number of novel points generated in the case of the measures L3 and N4.
While these methods have shown high performance for small nonimage datasets that are linearly separable, they are less suited for analyzing large and complex image datasets~\cite{brun2018framework}.
Furthermore, when these methods were proposed, they were intended for two-class datasets, which, despite being generalized by some scholars as multiclass datasets, do not address the abovementioned issues~\cite{orriols2010documentation, lorena2019complex}.
\par
Other methods have been proposed besides these 12 descriptors.
For example, the paper~\cite{baumgartner2006data} proposed a complexity assessment method for two-class high-dimensional biomedical datasets.
However, the method cannot handle multiclass datasets and requires the decomposition of $N \times d$ where $N$ and $d$ are the total number of training samples and the dimension of data, respectively.  
Also, the proposed method~\cite{wang2005euclidean} measures the similarity between two images with Euclidean distance, which cannot be generalized well to large and complex image datasets.
Other proposed methods were based on constructed graphs from the dataset to measure the intra- and interclass relations~\cite{garcia2015effect, pascual2020revisiting}.
These methods require the analysis of high-dimensional matrices and are therefore memory intensive and time-consuming.
\par
A recent method is CSG~\cite{branchaud2019spectral}, which has shown state-of-the-art dataset complexity assessment performance. 
It calculates dataset complexity based on the eigenvalues of a Laplacian matrix derived from the similarity matrix between the classes.
While it focuses on the gradient between adjacent eigenvalues ($i.e.$, eigengap) of the Laplacian matrix, it does not completely account for the influence of the AULS, which can also affect the spectrum's size.
\section{Proposed method}
\begin{figure}[p]
\begin{adjustbox}{addcode={
\begin{minipage}{\width}}{
\caption{Overview of the proposed method. It consists of three steps: (1) dimension reduction, (2) similarity matrix construction, and 3) dataset complexity calculation.}
\label{fig1}
\end{minipage}},rotate=90,center}
\includegraphics[width=18cm]{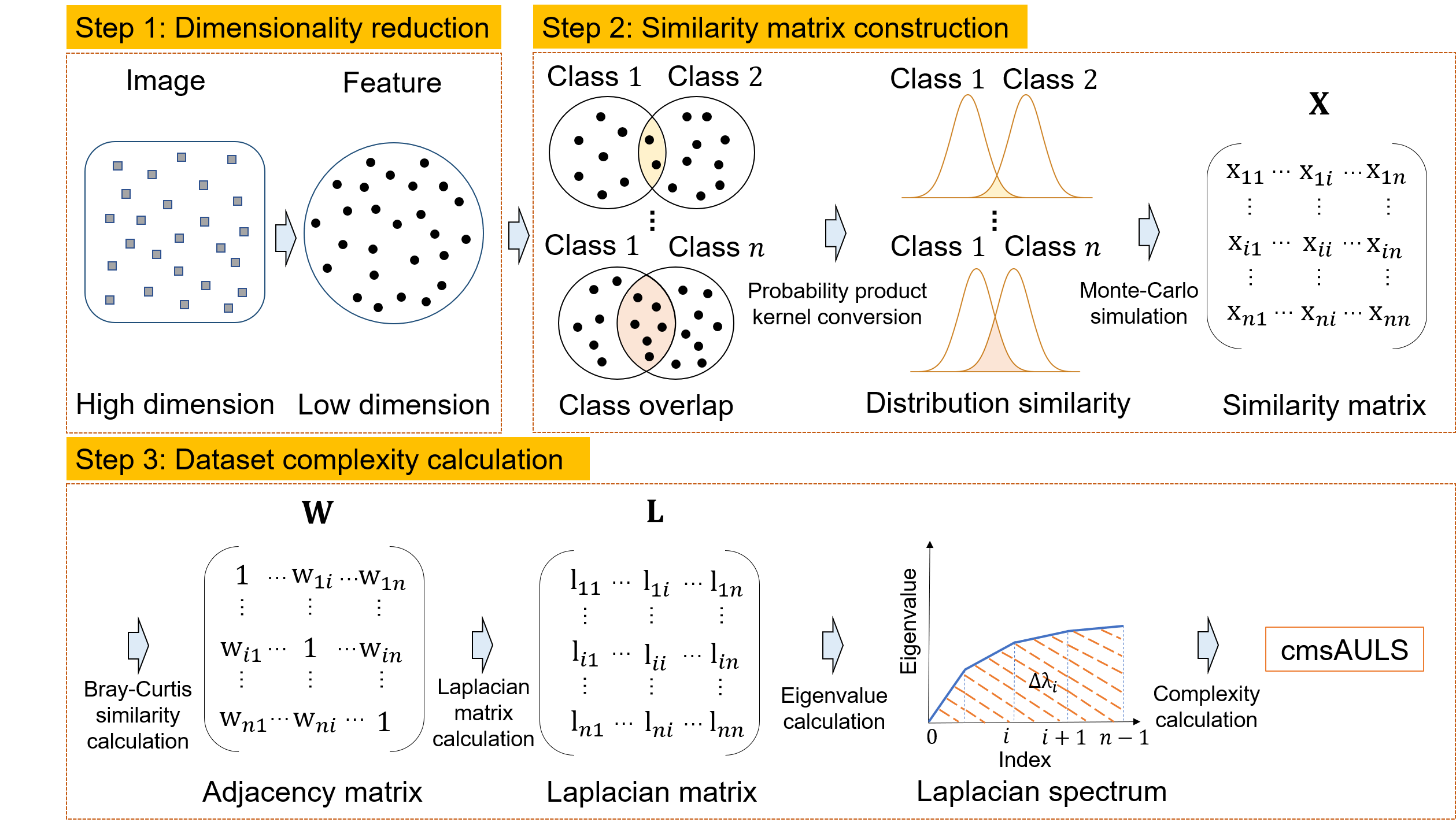}
\end{adjustbox}
\end{figure}
This section provides details of the proposed method, an overview of which is shown in Figure~\ref{fig1}.
In section 3.1, we describe the dimension reduction phase.
Then, in section 3.2, we demonstrate how to construct the similarity matrix between classes in a dataset.
Next, in section 3.3, we show the relation between spectral clustering and dataset complexity.
Finally, in section 3.4, we show how to calculate dataset complexity.
\subsection{Dimension reduction}
Since image data are generally high dimensional, they must be transformed into a new low-dimensional space by maintaining their characteristics.
Let $x$ be an input data, and the embedding of $x$ is defined as $\psi(x)\in\mathbb{R}^{d}$, where $d$ is the dimension of the downscaled feature.
$\psi$ can be any dimension reduction method ($e.g.$, autoencoder~\cite{wang2014generalized}, t-SNE~\cite{maaten2008visualizing}, and PCA~\cite{wold1987principal}).
${\psi}(x)$ is used to calculate class overlap in the next step.
\subsection{Similarity matrix construction}
The overlap between classes can denote an image dataset's complexity for classification problems~\cite{branchaud2019spectral}.
Therefore, the proposed method calculates dataset complexity based on the overlap between classes.
Although there are $n$ classes in a dataset, we can take two of them and calculate the overlap for the entire dataset.
From the integral measure of the Gaussian mixture model~\cite{nowakowska2014tractable}, when two classes $\mathcal{A}$ and $\mathcal{B}$ exist, the overlap between $\mathcal{A}$ and $\mathcal{B}$ refers to the overall area in the image feature space for which $P({\psi}(x_{t}) \mid \mathcal{B}) > P({\psi}(x_{t}) \mid \mathcal{A})$ when ${\psi}(x_{t})$ belongs to class $\mathcal{A}$. 
Hence, we can define the class overlap as follows:
\begin{equation}
\label{equ1}
\int_{\mathbb{R}^{d}} \mathrm{min} ( P({\psi}(x) \mid \mathcal{A}), P({\psi}(x) \mid \mathcal{B}))  \, d{\psi}(x),
\end{equation}
where $P({\psi}(x) \mid \mathcal{A})$ and $P({\psi}(x) \mid \mathcal{B})$ denote the distribution of the image feature ${\psi}(x)$ belonging to class $\mathcal{A}$ and $\mathcal{B}$, respectively.
Since calculating the integral directly is prohibitively complicated, based on the strong correlation between class overlap and similarity in data distribution, we can use the probability product kernel~\cite{jebara2004probability} to surrogate Eq. (1) as follows:
\begin{equation}
\label{equ2}
\int_{\mathbb{R}^{d}} P({\psi}(x) \mid \mathcal{A})^{\rho} \, P({\psi}(x) \mid \mathcal{B})^{\rho} \,   d{\psi}(x).
\end{equation}
When $\rho = 1$, the inner product between the two distributions is the expectation of one distribution under the other ($i.e.$, $\mathbb{E}_{P({\psi}(x) \mid \mathcal{A})}[P({\psi}(x) \mid \mathcal{B})]$ or $\mathbb{E}_{P({\psi}(x) \mid \mathcal{B})}[P({\psi}(x) \mid \mathcal{A})]$).
Since classes $\mathcal{A}$ and $\mathcal{B}$ have many images, directly calculating the expectation leads to inefficiency.
Therefore, we use the Monte Carlo method~\cite{binder1993monte} to approximate the expectation calculation process as follows:
\begin{equation}
\label{equ3}
\mathbb{E}_{P({\psi}(x) \mid \mathcal{A})}[P({\psi}(x) \mid \mathcal{B})]
\approx \frac{1}{M}\sum_{m=1}^{M}p({\psi}(x_{m}) \mid \mathcal{B}), 
\end{equation}
where ${\psi}(x_{m})(m=1,2,\cdots,M)$ are $M$ samples randomly selected from class $\mathcal{A}$, and $p({\psi}(x_{m}) \mid \mathcal{B})$ denotes the probability of ${\psi}(x_{m})$ belonging to class $\mathcal{B}$.
We can calculate the expectation between all classes and construct the similarity matrix $\mathbf{X}\in\mathbb{R}^{n \times n}$.
In addition, we use a $k$-nearest estimator to approximate $p({\psi}(x_{m}) \mid \mathcal{B})$ as follows:
\begin{equation}
\label{equ4}
p({\psi}(x_{m}) \mid \mathcal{B}) = \frac{K}{EV},
\end{equation}
where $K$ denotes the number of neighbors of ${\psi}(x_{m})$ in class $\mathcal{B}$.
Also, $E$ and $V$ denote the number of samples randomly selected from class $\mathcal{B}$ and the volume of the hypercube consisting of $k$ closest neighbors around ${\psi}(x_{m})$ in class $\mathcal{B}$, respectively.
\subsection{Spectral clustering}
In this section, we show the relation between spectral clustering and dataset complexity.
The calculated similarity matrix $\mathbf{X}$ contains the complexity information of a whole dataset, and we must derive a metric from $\mathbf{X}$ based on spectral clustering theory~\cite{von2007tutorial}. 
Let $G$ be an undirected similarity graph with nodes and edges.
The weight ($w_{ij} \geq 0$) of an edge that connects two nodes $i$ and $j$ denote their proximity. 
The weights of all edges are put in an $n \times n$ adjacency matrix $\mathbf{W}$, where $n$ is the total number of nodes.
The goal of spectral clustering is to partition $G$ into a set of subgraphs $\{G_{1}, \cdots, G_{i}, \cdots, G_{j}, \cdots, G_{r}\}$ to make the edges between subgraphs have minimum weight, where $G_{i} \cap G_{j} = \varnothing$, $\forall i \neq j$ and $ G_{1} \cup \cdots, \cup \,G_{r} = G$.
The optimal partition of $G$ needs to ensure that the cut is at a minimum cost: $\mathrm{Cut}\,({G_{1}, \cdots, G_{r}}) = \sum w_{ij}$ for $i$ and $j$ in different subgraphs.
\par
Spectral clustering provides a method to solve the partition problem via the Laplacian spectrum.
We first construct the Laplacian matrix $\mathbf{L}$ with the adjacency matrix $\mathbf{W}$ and the degree matrix $\mathbf{D}$ as follows:
\begin{equation}
\label{equ6}
\mathbf{L} = \mathbf{D} - \mathbf{W},
\end{equation}
\begin{equation}
\label{equ7}
\mathbf{D}_{ii} = \sum_{j = 1}^{n} \mathbf{W}_{ij}.
\end{equation}
The spectrum of the Laplacian matrix $\mathbf{L}$ contains $n$ eigenvalues $\lambda_{0}$, $\lambda_{1}$, $\cdots$, $\lambda_{n-1}$ $(\lambda_{0} = 0$, and $\lambda_{i+1} > \lambda_{i})$.
The $n$ eigenvectors associated with the eigenvalues can be seen as indicator vectors that one can use to cut the graph.
Also, the magnitude of their associated eigenvalues is related to the cost of their cut~\cite{mohar1997some}.
Therefore, the eigenvectors associated with the lowest eigenvalues are those associated with partitions of minimum cost.
\par
We can transfer the dataset complexity assessment problem into the spectral clustering framework with each node as a dataset class index.
$\mathbf{W}$ and $\mathbf{L}$ are $n \times n$ matrices where $n$ is the total number of classes. 
The weight $w_{ij}$ is the similarity between different classes.
Hence, a complex dataset with a large class overlap can lead to a Laplacian spectrum with large eigenvalues.  
The Laplacian spectrum magnitude expresses the similarity between classes and can be used to calculate a dataset's complexity.
\subsection{Dataset complexity calculation}
Since the similarity matrix $\mathbf{X}$ derived from the Monte Carlo method is not symmetric, it cannot be used as the adjacency matrix for calculating the Laplacian matrix $\mathbf{L}$.
Hecne, we first convert the similarity matrix $\mathbf{X}$ to a symmetric similarity matrix $\mathbf{W}\in\mathbb{R}^{n \times n}$ via the Bray Curtis distance~\cite{beals1984bray}:
\begin{equation}
\label{equ5}
\mathbf{W}_{ij} = 1 - \frac{\sum_{q=1}^{q=n}|\mathbf{X}_{iq} - \mathbf{X}_{jq}|}{\sum_{q=1}^{q=n}|\mathbf{X}_{iq} + \mathbf{X}_{jq}|},
\end{equation}
where $\mathbf{X}_{i}$ and $\mathbf{X}_{j}$ are the columns of the similarity matrix $\mathbf{X}$.
$\mathbf{W}_{ij}$ denotes the similarity between class $i$ and class $j$.
We can then construct the Laplacian matrix $\mathbf{L}$ with the symmetric adjacency matrix $\mathbf{W}$ and the degree matrix $\mathbf{D}$. 
The spectrum of the Laplacian matrix $\mathbf{L}$ contains $n$ eigenvalues $\lambda_{0}$, $\lambda_{1}$, $\cdots$, $\lambda_{n-1}$ $(\lambda_{0} = 0$ and $\lambda_{i+1} > \lambda_{i})$.
Considering that both the AULS and the gradient between adjacent eigenvalues can affect assessment performance, we propose cmsAULS, which is a simple but effective method for evaluating dataset complexity:
\begin{equation}
\label{equ8}
\mathrm{cmsAULS} = \sum_{i = 0}^{n - 2} \mathrm{cummax}(\Delta\lambda)_{i},
\end{equation}
\begin{equation}
\label{equ9}
\Delta\lambda_{i} = \frac{\lambda_{i+1} - \lambda_{i} }{n - i} \times \frac{\lambda_{i+1} + \lambda_{i} }{2} = \frac{\lambda_{i+1}^{2} - \lambda_{i}^{2} }{2(n - i)},
\end{equation}
where the $\mathrm{cummax}$ denotes the cumulative maximum value of a vector.
A small cmsAULS value indicates that the dataset has a small overlap between classes and vice versa.
Since the complexity calculation of cmsAULS is only related to the $n \times n$ size matrix calculation, the asymptotic time complexity of cmsAULS is $O(M \cdot d^{2} \cdot n^{2})$, where the number of selected samples $M$ and downscaled dimension $d$ are definite.
\begin{figure}[t]
        \centering
        \subfigure[]{
        \centering
        \includegraphics[width=5.5cm]{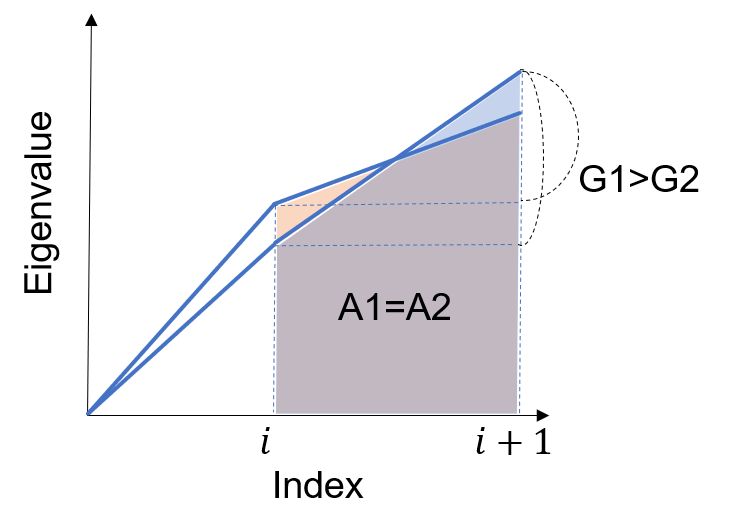}
        }
        \subfigure[]{
        \centering
        \includegraphics[width=5.5cm]{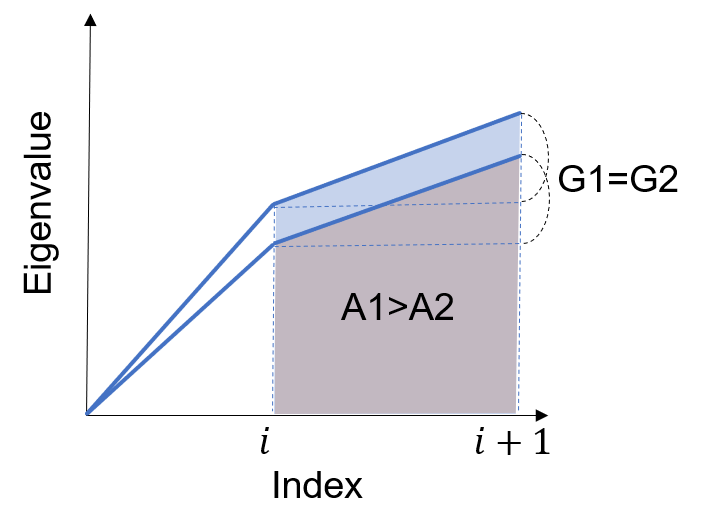}
        }
        \caption{The size of the Laplacian spectrum can be affected by two elements: (a) the gradient between adjacent eigenvalues and (b) the Area Under Laplacian Spectrum.}
        \label{fig2}
\end{figure}
\par
Figure~\ref{fig2} shows the concept illustration for cmsAULS.
If the extreme situation in Figure~\ref{fig2}-(a) occurs, ($i.e.$, two datasets with the same AULS), the gradient of the adjacent eigenvalues should be used to evaluate dataset complexity.
Meanwhile, the AULS is more suitable for assessing dataset complexity when there are two datasets with an equal gradient between specific adjacent eigenvalues as shown in Figure~\ref{fig2}-(b).
The cmsAULS focuses on both the gradient between adjacent eigenvalues and the AULS, which can achieve better assessment performance.
The proposed method is summarized in Algorithm~\ref{alg1}.
\begin{algorithm}[t]
    \caption{The proposed method}    
    \label{alg1}
    \begin{algorithmic}[1]
    \REQUIRE 
    a dataset with $n$ classes, $M$, $E$, $k$  
    \ENSURE
    cmsAULS 
    \\
    \STATE
    Reduce the dimensions of image data in the input dataset
    \STATE
    Compute similarity matrix $\mathbf{X}$ of the input dataset with Eqs.~\ref{equ3} and~\ref{equ4}
    \STATE
    Convert the similarity matrix $\mathbf{X}$ to a symmetric similarity matrix $\mathbf{W}$ with Eq.~\ref{equ5}
    \STATE
    Construct the Laplacian matrix $\mathbf{L}$ with Eqs.~\ref{equ6} and~\ref{equ7}
    \STATE
    $\{\lambda_{0}$, $\lambda_{1}$, $\cdots$, $\lambda_{n-1}\}$ $\leftarrow$ Eigenvalues\,($\mathbf{L}$)
    \STATE
    Compute cmsAULS with Eqs.~\ref{equ8} and~\ref{equ9}
    \end{algorithmic}
\end{algorithm}
\section{Experiments}
In this section, we conducted several experiments to verify the effectiveness of the cmsAULS. 
In section 4.1, we show the datasets used in our experiments.
Afterward, in section 4.2, we compare cmsAULS with several benchmark and state-of-the-art methods.
Then, in section 4.3, we test pretrained DCNN feature extractors combined with cmsAULS for a higher Pearson correlation.
Next, in section 4.4, we visualize the interclass distance of different datasets to verify the effectiveness of the obtained similarity matrix.
Finally, in section 4.5, we show the influence of different reduced dimensions for cmsAULS.
\subsection{Datasets}
To evaluate the performance of cmsAULS, we used six types of 10-class image classification datasets with various complexity levels, similar to those in~\cite{branchaud2019spectral}.
These datasets contain the well-known mnist~\cite{lecun2010mnist}, svhn~\cite{netzer2011reading} and cifar10~\cite{krizhevsky2009learning}.
NotMNIST~\cite{bulatov2011notmnist} is a dataset similar to mnist but consists of alphabets extracted from some publicly available fonts.
Also, stl10~\cite{coates2011analysis} is a cifar10-inspired dataset with each class having fewer labeled training examples than in cifar10 and with larger images ($i.e.$, 96 $\times$ 96).
Finally, compcars~\cite{yang2015large} is a dataset containing 163 car makes with 1,716 car models.
In our experiments, we selected the 10 highest counts of makes and resized the images to 128 $\times$ 128, providing 500 samples per class.
\subsection{Comparison with benchmark and the state-of-the-art methods}
\begin{table}[]
\begin{adjustbox}{addcode={
\begin{minipage}{\width}}{
\caption{Pearson correlation and p-value between the complexity and the test error rates of the six 10-class datasets. CAE denotes a 9-layer CNN autoencoder, and Comb. denotes the combination of CNN autoencoder and t-SNE.}
\label{tab2}
\end{minipage}},rotate=90,center}
    \begin{tabular}{c|ccc|ccc|ccc}
    \hline
    & & AlexNet & & & ResNet50 & & & Xception\\\hline
    Method & CAE & t-SNE & Comb. & CAE & t-SNE & Comb. & CAE & t-SNE & Comb. \\\hline
    F1
    & -0.575 (0.233) & -0.485 (0.329) & -0.522 (0.288)
    & -0.582 (0.225) & -0.458 (0.361) & -0.469 (0.348) 
    & -0.543 (0.266) & -0.413 (0.416) & -0.440 (0.382) \\
    F2 
    & -0.357 (0.487) & -0.061 (0.908) & 0.232 (0.659)
    & -0.370 (0.470) & -0.024 (0.964) & 0.158 (0.765) 
    & -0.317 (0.541) & -0.093 (0.862) & 0.151 (0.775) \\
    F3
    & -0.461 (0.357) & -0.262 (0.616) & -0.423 (0.403)
    & -0.424 (0.402) & -0.287 (0.582) & -0.365 (0.476) 
    & -0.375 (0.464) & -0.207 (0.694) & -0.328 (0.526) \\
    F4 
    & 0.229 (0.663) & -0.335 (0.516) & -0.417 (0.411)
    & 0.186 (0.725) & -0.333 (0.519) & -0.356 (0.488)
    & 0.276 (0.597) & -0.266 (0.610) & -0.317 (0.541) \\
    N1
    & 0.771 (0.073) & 0.704 (0.119) & 0.710 (0.114)
    & 0.712 (0.112) & 0.663 (0.151) & 0.653 (0.160) 
    & 0.677 (0.140) & 0.612 (0.196) & 0.613 (0.196) \\
    N2 
    & 0.683 (0.135) & 0.688 (0.131) & 0.778 (0.068)
    & 0.634 (0.177) & 0.647 (0.165) & 0.680 (0.137) 
    & 0.590 (0.218) & 0.592 (0.216) & 0.667 (0.148) \\
    N3
    & 0.776 (0.070) & 0.741 (0.092) & 0.744 (0.090)
    & 0.709 (0.115) & 0.692 (0.127) & 0.666 (0.148) 
    & 0.676 (0.140) & 0.644 (0.167) & 0.637 (0.174) \\
    N4 
    & 0.269 (0.606) & 0.552 (0.256) & 0.581 (0.227)
    & 0.133 (0.802) & 0.525 (0.285) & 0.537 (0.272) 
    & 0.130 (0.806) & 0.473 (0.344) & 0.497 (0.316) \\
    T1
    & 0.418 (0.410) & -0.563 (0.244) & 0.209 (0.691)
    & 0.356 (0.488) & -0.720 (0.107) & 0.092 (0.862) 
    & 0.341 (0.508) & -0.635 (0.175) & 0.101 (0.849) \\
    T2 
    & -0.774 (0.071) & -0.774 (0.071) & -0.774 (0.071)
    & -0.746 (0.088) & -0.746 (0.088) & -0.746 (0.088)
    & -0.785 (0.065) & -0.785 (0.065) & -0.785 (0.065) \\\hline
    AULS
    & 0.722 (0.105) & 0.921 (0.009) & 0.933 (0.006)
    & 0.722 (0.105) & 0.911 (0.012) & 0.907 (0.012) 
    & 0.665 (0.150) & 0.895 (0.016) & 0.888 (0.018) \\
    CSG 
    & 0.690 (0.129) & 0.908 (0.012) & 0.900 (0.015)
    & 0.740 (0.093) & 0.938 (0.006) & 0.943 (0.005) 
    & 0.676 (0.140) & 0.914 (0.011) & 0.911 (0.011) \\
    \bfseries{cmsAULS} 
    & 0.753 (0.084) & 0.962 (0.002) & \bfseries{0.969 (0.001)}
    & 0.838 (0.037) & 0.955 (0.003) & \bfseries{0.961 (0.002)}
    & 0.784 (0.065) & 0.941 (0.005) & \bfseries{0.950 (0.004)} \\
    \hline
    \end{tabular}
\end{adjustbox}
\end{table}
\begin{table}[t]
    \centering
    \caption{The network structure of the 9-layer CNN autoencoder.}
    \label{tab3}
    \begin{tabular}{c|c|c|c}
    \hline
    Layers & Operator & Resolution & Channels \\\hline
    1 & Conv3 $\&$ MaxPool & 32 $\times$ 32 & 64 \\
    2 & Conv3 $\&$ MaxPool & 16 $\times$ 16 & 128 \\
    3 & Conv3 $\&$ MaxPool & 8 $\times$ 8 & 256 \\
    4 & Conv3 $\&$ MaxPool & 4 $\times$ 4 & 256 \\
    5 & Conv1 & 4 $\times$ 4 & 8 \\
    6 & TConv2 & 8 $\times$ 8 & 128 \\
    7 & TConv2 & 16 $\times$ 16 & 256 \\
    8 & TConv2 & 32 $\times$ 32 & 512 \\
    9 & TConv2 & 64 $\times$ 64 & 512 \\
    \hline
    \end{tabular}
\end{table}
In this section, we compare cmsAULS with several benchmark and the state-of-the-art techniques.
In the dimension reduction phase, we use different methods (CNN autoencoder, t-SNE, and their combination) validate of our method.
The dimension of the downscaled image feature $d$ by CNN autoencoder and t-SNE are set to 128 and 3, respectively.
Similar to the paper, we set the hyperparameters $M$, $E$, and $k$ of the matrix construction phase to 100, 100 and 3, respectively, which can effectively calculate the complexity.
In the complexity calculation phase, we use 10 different descriptors~\cite{ho2002complexity, orriols2010documentation}, CSG~\cite{branchaud2019spectral} and the AULS as comparative methods for verifying the validity of cmsAULS. 
Finally, we use Pearson correlation and p-value between the error rates of three DCNN models (AlexNet~\cite{krizhevsky2012imagenet}, ResNet50~\cite{he2016deep}, and Xception~\cite{chollet2017xception}) and dataset complexity to evaluate the assessment performance of these methods. 
\par
\begin{figure}[t]
        \centering
        \includegraphics[width=9cm]{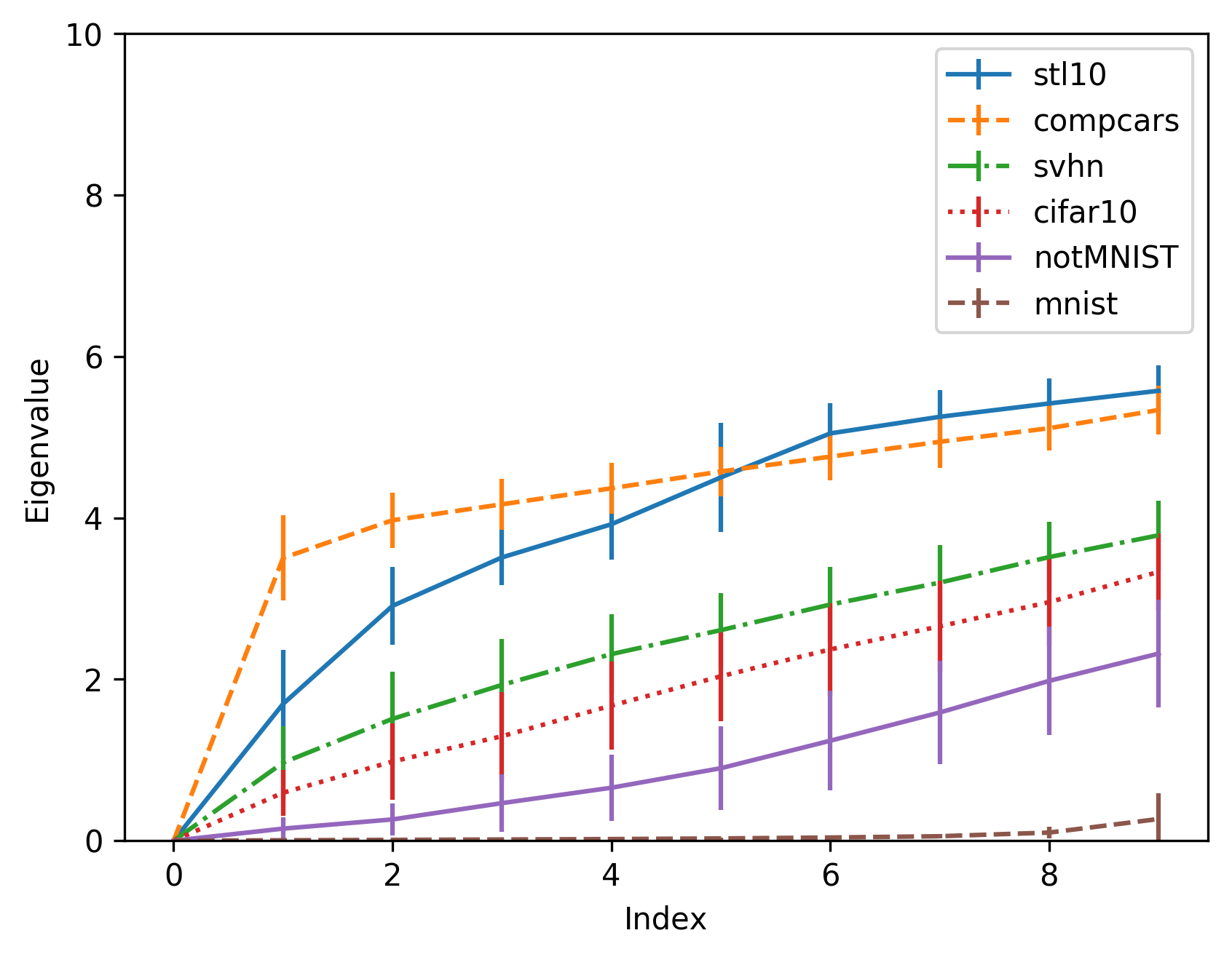}
        \caption{Laplacian spectrum of the six 10-class datasets (Comb.).}
        \label{fig3}
\end{figure}

Table~\ref{tab2} shows the Pearson correlation and p-value between the complexity and the test error rates of the six 10-class datasets. 
CAE denotes a 9-layer CNN autoencoder and Comb. pertains to the combination of CNN autoencoder and t-SNE.
The network structure of the 9-layer CNN autoencoder is shown in Table~\ref{tab3}, where Conv, MaxPool, and TConv denote convolution layer, maxpooling layer, and transposed convolution layer, respectively. 
The number behind Conv denotes the kernel size.
Table~\ref{tab2} shows that although the neighborhood methods N1, N2, N3, and N4 outperform other benchmark methods, they cannot even achieve a Pearson correlation of 0.8.
However, cmsAULS outperforms all other methods with a large margin with an average Pearson correlation of 0.96.
We can confirm that the complexity calculated by cmsAULS has a high Pearson correlation with DCNN test error rates based on the results in Table~\ref{tab2}.
Meanwhile, Table~\ref{tab4} shows the test error rate results for the three DCNN models on the six 10-class datasets.
For fairness, we use the test error rates results directly reported in~\cite{branchaud2019spectral}.
Table~\ref{tab5} shows the calculated complexity of the six 10-class datasets.
From Tables~\ref{tab2} and~\ref{tab5}, we can see that a simple dataset has a low complexity score ($e.g.$, mnist) and vice versa.
Figure~\ref{fig3} shows the Laplacian spectrum of the six 10-class datasets.
From Figure~\ref{fig3}, we can confirm that a dataset with high test error rates tends to have a large Laplacian spectrum.
\begin{table}[t]
    \centering
    \caption{Test error rates for three DCNN models on the six 10-class datasets~\cite{branchaud2019spectral}.}
    \label{tab4}
    \begin{tabular}{lccc}
    \\
    \hline
    Dataset & AlexNet & ResNet50 & Xception \\\hline\hline
    mnist & 0.01 & 0.05 & 0.01 \\
    notMNIST & 0.05 & 0.04 & 0.03 \\
    svhn & 0.08 & 0.07 & 0.03 \\
    cifar10 & 0.18 & 0.19 & 0.06 \\
    stl10 & 0.69 & 0.63 & 0.69 \\
    compcars & 0.70 & 0.88 & 0.86\\
    \hline
    \end{tabular}
\end{table}
\begin{table}[t]
    \centering
    \caption{Complexity of the six 10-class datasets (Comb.).}
    \label{tab5}
    \begin{tabular}{lccc}
    \\
    \hline
    Dataset & cmsAULS & CSG & AULS \\\hline\hline
    mnist & 0.144 & 0.045 & 0.675 \\
    notMNIST & 0.693 & 0.747 & 9.294 \\
    svhn & 1.100 & 1.826 & 20.142 \\
    cifar10 & 1.224 & 2.043 & 22.112 \\
    stl10 & 1.914 & 3.546 & 49.134 \\
    compcars & 3.170 & 3.840 & 58.353 \\
    \hline
    \end{tabular}
\end{table}
\subsection{The effectiveness of pretrained DCNN feature extractors}
\begin{figure}[t]
        \centering
        \includegraphics[width=9cm]{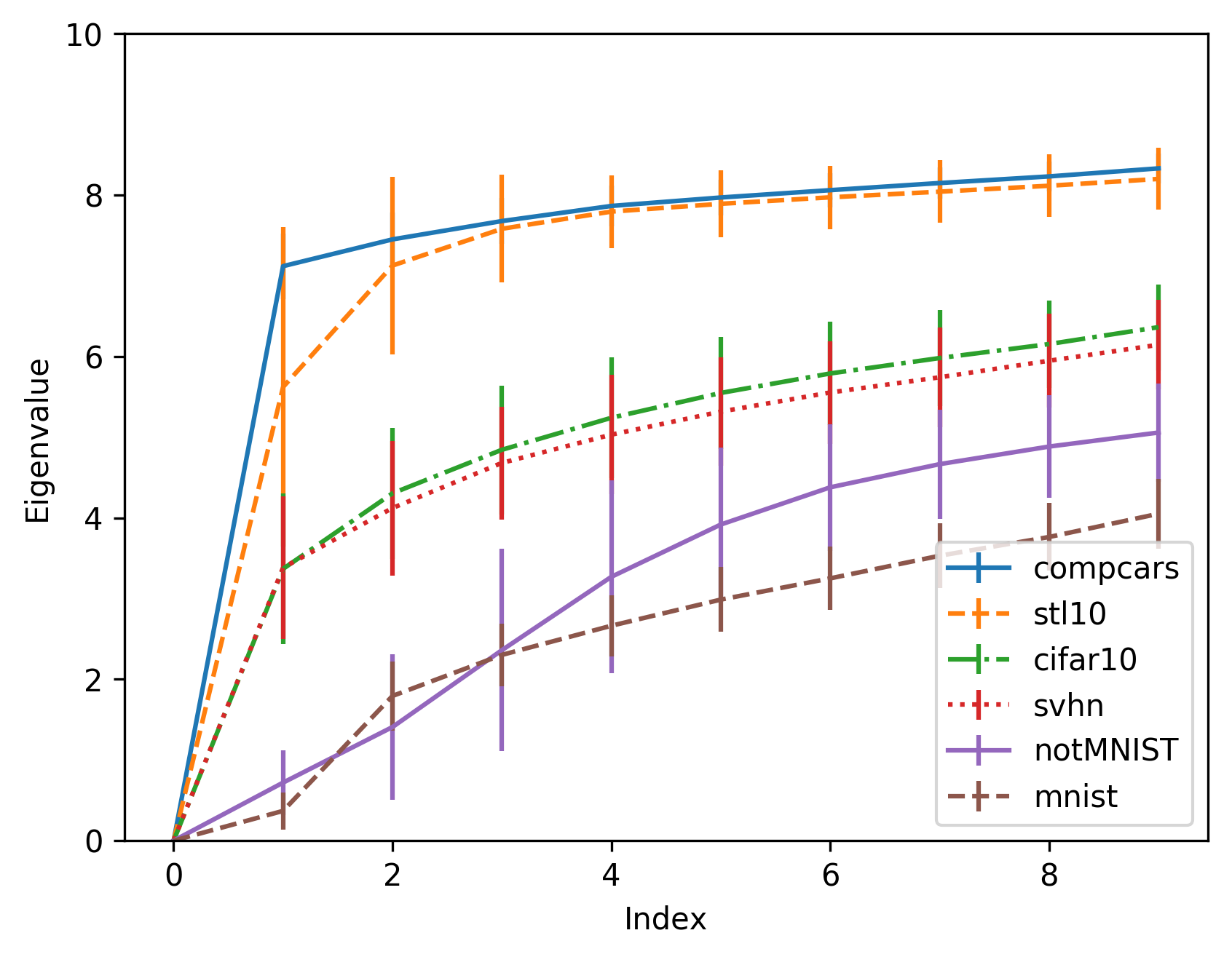}
        \caption{Laplacian spectrum of the six 10-class datasets (EfficientNet-B4 and t-SNE).}
        \label{fig4}
\end{figure}
\begin{table}[t]
    \centering
    \caption{Pearson correlation and p-value between the complexity and the test error rates of the six 10-class datasets.}
    \label{tab6}
    \begin{tabular}{lcccc}
    \\
    \hline
    Method & Evaluation & AlexNet & ResNet50 & Xception \\\hline\hline
    \bfseries{cmsAULS} & Corr & \bfseries{0.989} & \bfseries{0.986} & \bfseries{0.988} \\
    \bfseries{cmsAULS} & p-val & \bfseries{$<$0.001} & \bfseries{$<$0.001} & \bfseries{$<$0.001} \\\hline
    CSG & Corr & 0.956 & 0.965 & 0.948 \\
    CSG & p-val & 0.003 & 0.002 & 0.004 \\\hline
    AULS & Corr & 0.942 & 0.913 & 0.898 \\
    AULS & p-val & 0.005 & 0.011 & 0.015 \\
    \hline
    \end{tabular}
\end{table}
\begin{table}[t]
    \centering
    \caption{Pearson correlation between the complexity and the test error rates of the five 10-class datasets (excluding one of the six datasets).}
    \label{tab7}
    \begin{tabular}{lcccc}
    \\
    \hline
    Remove & Method & AlexNet & ResNet50 & Xception  \\\hline\hline
     & \bfseries{cmsAULS} & \bfseries{0.988} & \bfseries{0.992} & \bfseries{0.996} \\
    mnist & CSG & 0.952 & 0.978 & 0.960 \\
     & AULS & 0.951 & 0.936 & 0.922  \\
     \hline
     & \bfseries{cmsAULS} & \bfseries{0.988} & \bfseries{0.985} & \bfseries{0.987} \\
    notMNIST & CSG & 0.952 & 0.961 & 0.947 \\
     & AULS & 0.936 & 0.900 & 0.892 \\
    \hline
    & \bfseries{cmsAULS} & \bfseries{0.992} & \bfseries{0.991} & \bfseries{0.991} \\
    svhn & CSG & 0.976 & 0.989 & 0.968 \\
     & AULS & 0.975 & 0.949 & 0.929 \\
    \hline
    & \bfseries{cmsAULS} & \bfseries{0.989} & \bfseries{0.987} & \bfseries{0.991} \\
    cifar10 & CSG & 0.957 & 0.967 & 0.962 \\
     & AULS & 0.952 & 0.924 & 0.931 \\
    \hline
    & \bfseries{cmsAULS} & \bfseries{0.994} & \bfseries{0.988} & \bfseries{0.984} \\
    stl10 & CSG & 0.973 & 0.957 & 0.939 \\
     & AULS & 0.921 & 0.893 & 0.859 \\
    \hline
    & \bfseries{cmsAULS} & \bfseries{0.992} & \bfseries{0.980} & \bfseries{0.976} \\
    compcars & CSG & 0.937 & 0.924 & 0.887 \\
     & AULS & 0.908 & 0.894 & 0.845 \\
    \hline
    \end{tabular}
\end{table}
In this section, we test pretrained DCNN feature extractors combined with cmsAULS for a higher Pearson correlation.
Also, we calculate the Pearson correlation between the complexity and the test error rates of five 10-class datasets (one of the six datasets was removed) to verify the robustness of cmsAULS.
We use EfficientNet~\cite{tan2019efficientnet} trained with Noisy Student~\cite{xie2020self} as feature extractors, which are the most prominent image classification methods of ImageNet~\cite{deng2009imagenet}.
Specifically, we use EfficientNet-B4 extractors trained on ImageNet, which performance better than other EfficientNet versions in our experiments.
Furthermore, Noisy Student training is a semisupervised learning approach that works well even when labeled data are abundant, improving the classification performance of supervised learning.
Therefore, EfficientNet trained with Noisy Student tends to obtain a better feature representation of images.
Since the t-SNE method performed well in the above experiments, we use EfficientNet-B4 combined with t-SNE to reduce the dimension of the extracted image feature in this experiment.
\par
Table~\ref{tab6} shows the Pearson correlation and p-value between the complexity and the test error rates of the six 10-class image datasets.
From Table~\ref{tab6}, we can see that the proposed method has a better correlation with all three models than CSG and AULS.  
Furthermore, cmsAULS has the lowest p-value ($<$0.001), which indicates reliable assessment results.
Figure~\ref{fig4} shows the Laplacian spectrum of the six 10-class datasets, which uses the combination of EfficientNet-B4 and t-SNE for reducing image feature dimension.
From Figure~\ref{fig4}, as in Figure~\ref{fig3}, we can confirm that a dataset with high test error rates tends to have a large Laplacian spectrum.
Table~\ref{tab7} shows the Pearson correlation between the complexity and the test error rates of the five 10-class datasets (one of the six datasets was removed).
From Table~\ref{tab7}, we can see that cmsAULS has a robust performance in the dataset complexity assessment task.
We can confirm the validity and robustness of cmsAULS with the results in Tables~\ref{tab6} and~\ref{tab7}.
We think that the image feature extracted by EfficientNet-B4 is more similar to the tested DCNN models (AlexNet, ResNet50, and Xception), hence performing better than the CNN autoencoder.   
\subsection{Interclass distance visualization}
\begin{figure}[]
        \centering
        \subfigure[]{
        \centering
        \includegraphics[width=8.0cm]{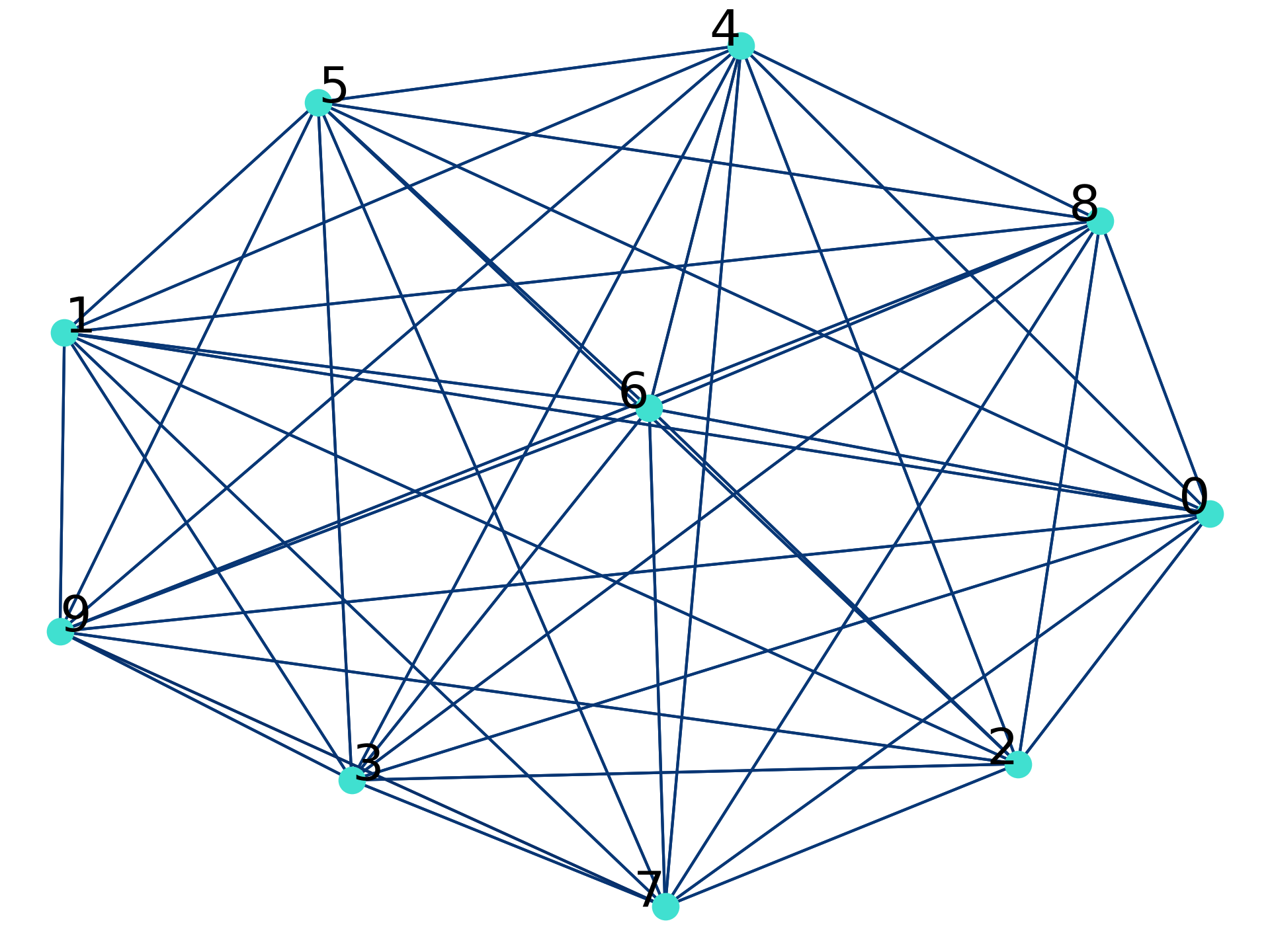}
        }
        \subfigure[]{
        \centering
        \includegraphics[width=8.0cm]{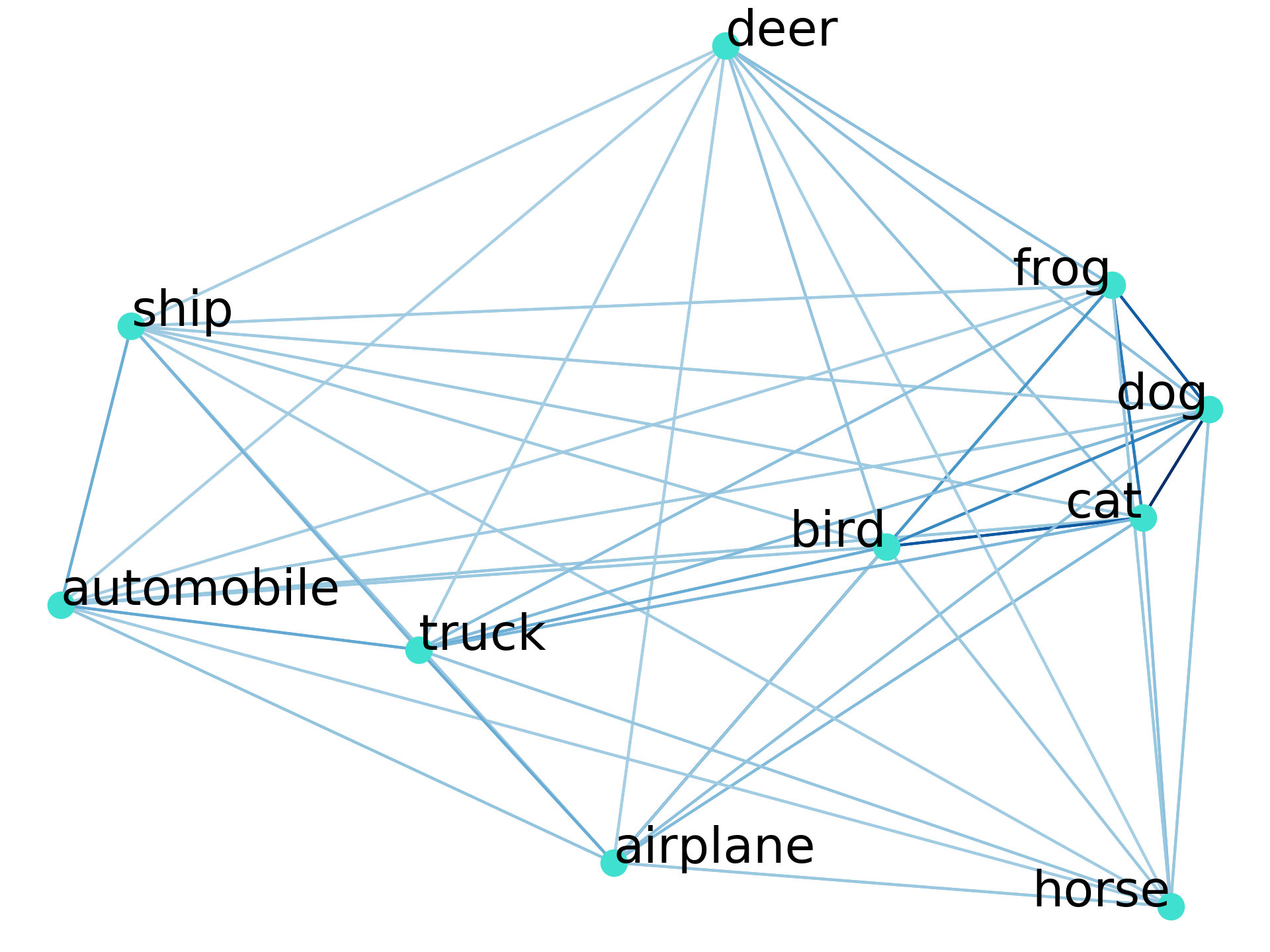}
        }
        \subfigure[]{
        \centering
        \includegraphics[width=8.0cm]{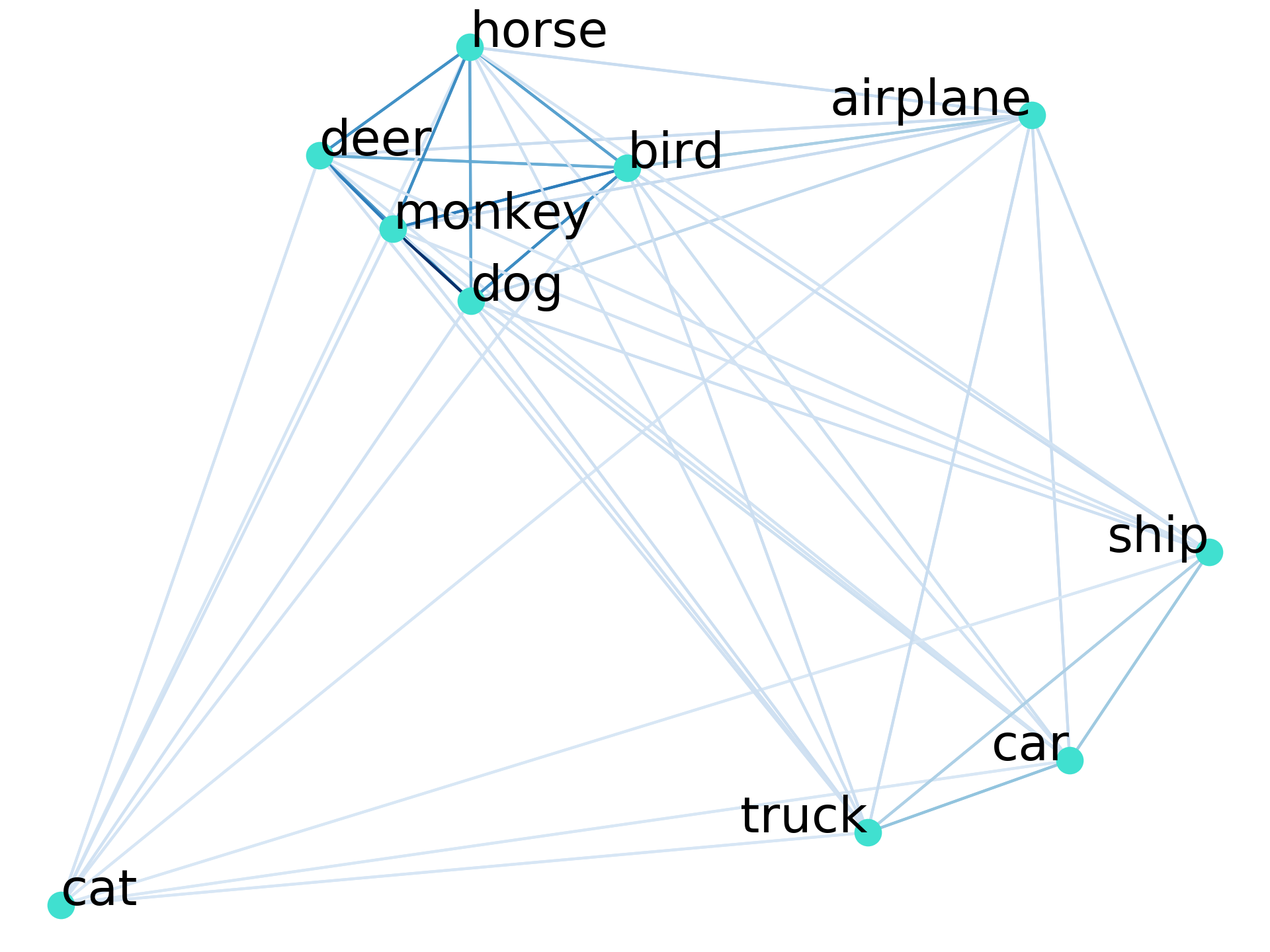}
        }
        \caption{Interclass distance of different datasets: (a) mnist, (b) cifar10, and (c) stl10.}
        \label{fig5}
\end{figure}
In this section, we visualize the interclass distance of different datasets to verify the effectiveness of the obtained similarity matrix.
Although we have proved that there is a high Pearson correlation between the dataset complexity calculated by our method and DCNN test error rates, we can also use the obtained similarity matrix $\mathbf{W}_{ij}$ to show the interclass distance in a dataset.
Specifically, we use the dissimilarity matrix $\mathbf{U}_{ij} = 1 - \mathbf{W}_{ij}$ to visualize the interclass distance of a dataset in two dimensions via multidimensional scaling (MDS)~\cite{borg2005modern}.
\par
Figure~\ref{fig5} shows the interclass distance of three datasets (mnist, cifar10, and stl10) with different complexity.
From Figure~\ref{fig5}, we can see that when a dataset has a high complexity, the distances between some classes are extremely small, showing that they are difficult to separate well.
For example, the classes of mnist are well separated, while the cat and dog classes in cifar10 as well as the deer and horse classes in stl10 are extremely close because of similar visual contents.
Therefore, we can also confirm that cifar10 and stl10 are more complex than mnist from the visualization results of interclass distance.
\subsection{The influence of the image feature's dimension}
In this section, we show the influence of different reduced dimensions for cmsAULS.
We use two prominent dimension reduction methods (t-SNE and PCA) with different reduced dimensions in the experiment. 
For t-SNE, we set the reduced dimension to 2 and 3, which are the most frequently used.
Also, we test PCA with the reduced dimension set to 3 and 50 and contribution rates of 0.90 and 0.95.
\par
Table~\ref{tab8} shows the experimental results.
We can see that when the reduced dimension is small ($e.g.$, three dimensions), t-SNE achieves the best performance and outperforms PCA by a large margin. 
When we set the reduced dimension with a contribution rate of 0.90, our method with PCA also achieves good performance with a faster running time. 
From Table~\ref{tab8}, we can confirm the selection of dimension reduction methods and the reduced dimension is important to our method.
\begin{table}[t]
    \centering
    \caption{Pearson correlation and p-value between the complexity and the test error rates of the six 10-class datasets with different reduced dimensions. 2$d$, 3$d$, and 50$d$ denote different reduced dimensions. 0.90 and 0.95 denote different PCA contribution rates.}
    \label{tab8}
    \begin{tabular}{lcccc}
    \\
    \hline
    Method & Evaluation & AlexNet & ResNet50 & Xception \\\hline\hline
    t-SNE (2$d$) & Corr & 0.511 & 0.402 & 0.401 \\
    t-SNE (2$d$) & p-val & 0.300 & 0.430 & 0.431 \\\hline
    t-SNE (3$d$)  & Corr & \bfseries{0.969} & \bfseries{0.961} & \bfseries{0.950} \\
    t-SNE (3$d$) & p-val & \bfseries{0.001} & \bfseries{0.002} & \bfseries{0.004} \\\hline
    PCA (3$d$) & Corr & 0.291 & 0.362 & 0.298 \\
    PCA (3$d$) & p-val & 0.575 & 0.481 & 0.567 \\\hline
    PCA (50$d$) & Corr & 0.784 & 0.877 & 0.813 \\
    PCA (50$d$) & p-val & 0.065 & 0.022 & 0.049 \\\hline
    PCA (0.90) & Corr & 0.796 & 0.887 & 0.825 \\
    PCA (0.90) & p-val & 0.058 & 0.019 & 0.043 \\\hline
    PCA (0.95) & Corr & 0.774 & 0.873 & 0.808 \\
    PCA (0.95) & p-val & 0.070 & 0.023 & 0.052 \\
    \hline
    \end{tabular}
\end{table}
\section{Discussion}
\begin{table}[t]
    \centering
    \caption{Time cost on cifar10 of different methods~\cite{branchaud2019spectral}.}
    \label{tab9}
    \begin{tabular}{lc}
    \\
    \hline
    Method & Time (s) \\\hline\hline
    F1 & 72 \\
    F2 & 72 \\
    F3 & 3,924 \\
    F4 & 3,644 \\
    N1 & 17,748 \\
    N2 & 36,180 \\
    N3 & 36,216 \\
    N4 & 3,744 \\
    T1 & 36,108 \\
    T2 & 72 \\\hline
    AULS & 50 \\
    CSG & 50 \\
    cmsAULS & 50 \\
    \hline
    \end{tabular}
\end{table}
In our experiments, we first compare cmsAULS with several benchmark and state-of-the-art methods to show its effectiveness.
Then, we test pretrained DCNN feature extractors combined with cmsAULS for a higher Pearson correlation.
We visualize the interclass distance of different datasets to verify the effectiveness of the obtained similarity matrix.
At last, we show the influence of different reduced dimensions for our method.
\par
In this study, we mainly verify the improvement in Pearson correlation between the complexity calculated by cmsAULS and DCNN test error rates.
Since in the paper~\cite{branchaud2019spectral}, the effectiveness of CSG in terms of running time compared with other benchmark methods has been verified, and our method has the same running time as CSG, we show the time cost on cifar10 of different methods in Table~\ref{tab9}.
Our method can apply to a large number class classification problem, but it will cost more time because it needs more samples $M$ and more classes $n$. 
On the other hand, it is hard for our method to generalize to the task for pixel-level classification problems such as segmentation because there are many different classes in one image, it will be one of our future works.
Furthermore, cmsAULS is not restricted to image data and could be applied to other multimedia data with specific embedding methods ($e.g.$, Word2Vec~\cite{mikolov2013distributed} and VGGish~\cite{hershey2017cnn}).
Moreover, our previous studies related to self-supervised learning~\cite{li2021cross, li2021self, li2021triplet} can learn discriminative representations from images without manually annotated labels, which fits well with dataset complexity assessment algorithms. 
\par
Our method also has limitations.
The proposed method uses the AULS and the area is associated with the maximum eigenvalue of the Laplacian matrix as with the Rayleigh quotient.
Therefore, the upper limit value of cmsAULS is uncertain and could be extremely large, although this does not affect performance.
CSG uses the normalization of eigengap and can therefore have an upper-limit value ($i.e.$, number of classes), which is a good feature that cmsAULS does not have.
\section{Conclusion}
In this paper, we propose a novel method called cmsAULS to improve assessment performance regarding image dataset complexity.
From spectral clustering theory, the Laplacian spectrum size can denote similarities between dataset classes and can therefore be used to assess dataset complexity.
Moreover, two elements can affect Laplacian spectrum size, the AULS and the gradient between adjacent eigenvalues.
These are the focus of our method, which achieves better assessment performance compared with previous methods.
As a result, our method outperforms state-of-the-art methods in dataset complexity assessment on six datasets.
\section*{Conflict of interest}
The authors declare that they have no conflict of interest.
\section*{Acknowledgements}
This work was partly supported by AMED Grant Number JP21zf0127004. This  study  was  conducted  on  the  Data  Science  Computing System of Education and Research Center for Mathematical and Data Science, Hokkaido University.
\bibliographystyle{spmpsci}
\bibliography{refs}

\begin{thebibliography}{10}
\providecommand{\url}[1]{{#1}}
\providecommand{\urlprefix}{URL }
\expandafter\ifx\csname urlstyle\endcsname\relax
  \providecommand{\doi}[1]{DOI~\discretionary{}{}{}#1}\else
  \providecommand{\doi}{DOI~\discretionary{}{}{}\begingroup
  \urlstyle{rm}\Url}\fi

\bibitem{anwar2014measurement}
Anwar, N., Jones, G., Ganesh, S.: Measurement of data complexity for
  classification problems with unbalanced data.
\newblock Statistical Analysis and Data Mining \textbf{7}(3), 194--211 (2014)

\bibitem{baumgartner2006data}
Baumgartner, R., Somorjai, R.L.: Data complexity assessment in undersampled
  classification of high-dimensional biomedical data.
\newblock Pattern Recognition Letters \textbf{27}(12), 1383--1389 (2006)

\bibitem{beals1984bray}
Beals, E.W.: Bray-curtis ordination: an effective strategy for analysis of
  multivariate ecological data.
\newblock Advances in ecological research \textbf{14}, 1--55 (1984)

\bibitem{binder1993monte}
Binder, K., Heermann, D., Roelofs, L., Mallinckrodt, A.J., McKay, S.: Monte
  carlo simulation in statistical physics.
\newblock Computers in Physics \textbf{7}(2), 156--157 (1993)

\bibitem{borg2005modern}
Borg, I., Groenen, P.J.: Modern multidimensional scaling: Theory and
  applications.
\newblock Springer Science \& Business Media (2005)

\bibitem{branchaud2019spectral}
Branchaud-Charron, F., Achkar, A., Jodoin, P.M.: Spectral metric for dataset
  complexity assessment.
\newblock In: Proceedings of the IEEE/CVF Conference on Computer Vision and
  Pattern Recognition (CVPR), pp. 3215--3224 (2019)

\bibitem{brun2018framework}
Brun, A.L., Britto~Jr, A.S., Oliveira, L.S., Enembreck, F., Sabourin, R.: A
  framework for dynamic classifier selection oriented by the classification
  problem difficulty.
\newblock Pattern Recognition \textbf{76}, 175--190 (2018)

\bibitem{bulatov2011notmnist}
Bulatov, Y.: Notmnist dataset.
\newblock [Online]. Available:
  \url{http://yaroslavvb.blogspot.com/2011/09/notmnist-dataset.html}, 2011

\bibitem{chollet2017xception}
Chollet, F.: Xception: Deep learning with depthwise separable convolutions.
\newblock In: Proceedings of the IEEE/CVF Conference on Computer Vision and
  Pattern Recognition (CVPR), pp. 1251--1258 (2017)

\bibitem{coates2011analysis}
Coates, A., Ng, A., Lee, H.: An analysis of single-layer networks in
  unsupervised feature learning.
\newblock In: Proceedings of the International Conference on Artificial
  Intelligence and Statistics (AISTATS), pp. 215--223 (2011)

\bibitem{deng2009imagenet}
Deng, J., Dong, W., Socher, R., Li, L.J., Li, K., Fei-Fei, L.: Imagenet: A
  large-scale hierarchical image database.
\newblock In: Proceedings of the IEEE Conference on Computer Vision and Pattern
  Recognition (CVPR), pp. 248--255 (2009)

\bibitem{duin2006object}
Duin, R.P., P{\k{e}}kalska, E.: Object representation, sample size, and data
  set complexity.
\newblock Springer (2006)

\bibitem{gal2016dropout}
Gal, Y., Ghahramani, Z.: Dropout as a bayesian approximation: Representing
  model uncertainty in deep learning.
\newblock In: Proceedings of the International Conference on Machine Learning
  (ICML), pp. 1050--1059 (2016)

\bibitem{garcia2015effect}
Garcia, L.P., de~Carvalho, A.C., Lorena, A.C.: Effect of label noise in the
  complexity of classification problems.
\newblock Neurocomputing \textbf{160}, 108--119 (2015)

\bibitem{he2016deep}
He, K., Zhang, X., Ren, S., Sun, J.: Deep residual learning for image
  recognition.
\newblock In: Proceedings of the IEEE/CVF Conference on Computer Vision and
  Pattern Recognition (CVPR), pp. 770--778 (2016)

\bibitem{hershey2017cnn}
Hershey, S., Chaudhuri, S., Ellis, D.P., Gemmeke, J.F., Jansen, A., Moore,
  R.C., Plakal, M., Platt, D., Saurous, R.A., Seybold, B., et~al.: Cnn
  architectures for large-scale audio classification.
\newblock In: Proceedings of the IEEE International Conference on Acoustics,
  Speech and Signal Processing (ICASSP), pp. 131--135 (2017)

\bibitem{ho2002complexity}
Ho, T.K., Basu, M.: Complexity measures of supervised classification problems.
\newblock IEEE Transactions on Pattern Analysis and Machine Intelligence
  \textbf{24}(3), 289--300 (2002)

\bibitem{hoiem2012diagnosing}
Hoiem, D., Chodpathumwan, Y., Dai, Q.: Diagnosing error in object detectors.
\newblock In: Proceedings of the IEEE European Conference on Computer Vision
  (ECCV), pp. 340--353 (2012)

\bibitem{jebara2004probability}
Jebara, T., Kondor, R., Howard, A.: Probability product kernels.
\newblock Journal of Machine Learning Research \textbf{5}, 819--844 (2004)

\bibitem{krizhevsky2009learning}
Krizhevsky, A., Hinton, G., et~al.: Learning multiple layers of features from
  tiny images  (2009)

\bibitem{krizhevsky2012imagenet}
Krizhevsky, A., Sutskever, I., Hinton, G.E.: Imagenet classification with deep
  convolutional neural networks.
\newblock In: Proceedings of the Advances in Neural Information Processing
  Systems (NeurIPS), pp. 1097--1105 (2012)

\bibitem{lecun2010mnist}
LeCun, Y., Cortes, C., Burges, C.: Mnist handwritten digit database.
\newblock [Online]. Available: \url{http://yann.lecun.com/exdb/mnist/}, 2010

\bibitem{leyva2014set}
Leyva, E., Gonz{\'a}lez, A., Perez, R.: A set of complexity measures designed
  for applying meta-learning to instance selection.
\newblock IEEE Transactions on Knowledge and Data Engineering \textbf{27}(2),
  354--367 (2014)

\bibitem{li2020complexity}
Li, G., Togo, R., Ogawa, T., Haseyama, M.: Complexity evaluation of medical
  image data for classification problem based on spectral clustering.
\newblock In: Proceedings of the IEEE Global Conference on Consumer Electronics
  (GCCE), pp. 667--669 (2020)

\bibitem{li2021cross}
Li, G., Togo, R., Ogawa, T., Haseyama, M.: Cross-view self-supervised learning
  via momentum statistics in batch normalization.
\newblock In: Proceedings of the IEEE International Conference on Consumer
  Electronics – Taiwan (ICCE-TW) (2021)

\bibitem{li2021self}
Li, G., Togo, R., Ogawa, T., Haseyama, M.: Self-supervised learning for
  gastritis detection with gastric x-ray images.
\newblock arXiv:2104.02864  (2021)

\bibitem{li2021triplet}
Li, G., Togo, R., Ogawa, T., Haseyama, M.: Triplet self-supervised learning for
  gastritis detection with scarce annotations.
\newblock In: Proceedings of the IEEE Global Conference on Consumer Electronics
  (GCCE) (2021)

\bibitem{liu2019rethinking}
Liu, Z., Sun, M., Zhou, T., Huang, G., Darrell, T.: Rethinking the value of
  network pruning.
\newblock In: Proceedings of the International Conference on Learning
  Representations (ICLR) (2019)

\bibitem{lorena2019complex}
Lorena, A.C., Garcia, L.P., Lehmann, J., Souto, M.C., Ho, T.K.: How complex is
  your classification problem? a survey on measuring classification complexity.
\newblock ACM Computing Surveys \textbf{52}(5), 1--34 (2019)

\bibitem{maaten2008visualizing}
Maaten, L.v.d., Hinton, G.: Visualizing data using t-sne.
\newblock Journal of Machine Learning Research \textbf{9}, 2579--2605 (2008)

\bibitem{mikolov2013distributed}
Mikolov, T., Sutskever, I., Chen, K., Corrado, G., Dean, J.: Distributed
  representations of words and phrases and their compositionality.
\newblock In: Proceedings of the Advances in Neural Information Processing
  Systems (NeurIPS) (2013)

\bibitem{mohar1997some}
Mohar, B.: Some applications of laplace eigenvalues of graphs.
\newblock In: Graph symmetry, pp. 225--275. Springer (1997)

\bibitem{netzer2011reading}
Netzer, Y., Wang, T., Coates, A., Bissacco, A., Wu, B., Ng, A.Y.: Reading
  digits in natural images with unsupervised feature learning.
\newblock In: Proceedings of the Advances in Neural Information Processing
  Systems (NeurIPS), Workshop (2011)

\bibitem{nowakowska2014tractable}
Nowakowska, E., Koronacki, J., Lipovetsky, S.: Tractable measure of component
  overlap for gaussian mixture models.
\newblock arXiv preprint arXiv:1407.7172  (2014)

\bibitem{orriols2010documentation}
Orriols-Puig, A., Macia, N., Ho, T.K.: Documentation for the data complexity
  library in c++.
\newblock Universitat Ramon Llull, La Salle \textbf{196}, 1--40 (2010)

\bibitem{pascual2020revisiting}
Pascual-Triana, J.D., Charte, D., Arroyo, M.A., Fern{\'a}ndez, A., Herrera, F.:
  Revisiting data complexity metrics based on morphology for overlap and
  imbalance: Snapshot, new overlap number of balls metrics and singular
  problems prospect.
\newblock arXiv preprint arXiv:2007.07935  (2020)

\bibitem{tan2019efficientnet}
Tan, M., Le, Q.V.: Efficientnet: Rethinking model scaling for convolutional
  neural networks.
\newblock In: Proceedings of the International Conference on Machine Learning
  (ICML), pp. 6105--6114 (2019)

\bibitem{von2007tutorial}
Von~Luxburg, U.: A tutorial on spectral clustering.
\newblock Statistics and computing \textbf{17}(4), 395--416 (2007)

\bibitem{wang2005euclidean}
Wang, L., Zhang, Y., Feng, J.: On the euclidean distance of images.
\newblock IEEE Transactions on Pattern Analysis and Machine Intelligence
  \textbf{27}(8), 1334--1339 (2005)

\bibitem{wang2014generalized}
Wang, W., Huang, Y., Wang, Y., Wang, L.: Generalized autoencoder: A neural
  network framework for dimensionality reduction.
\newblock In: Proceedings of the IEEE/CVF Conference on Computer Vision and
  Pattern Recognition (CVPR), Workshop, pp. 490--497 (2014)

\bibitem{wold1987principal}
Wold, S., Esbensen, K., Geladi, P.: Principal component analysis.
\newblock Chemometrics and intelligent laboratory systems \textbf{2}(1-3),
  37--52 (1987)

\bibitem{xie2020self}
Xie, Q., Luong, M.T., Hovy, E., Le, Q.V.: Self-training with noisy student
  improves imagenet classification.
\newblock In: Proceedings of the IEEE/CVF Conference on Computer Vision and
  Pattern Recognition (CVPR), pp. 10687--10698 (2020)

\bibitem{yang2015large}
Yang, L., Luo, P., Change~Loy, C., Tang, X.: A large-scale car dataset for
  fine-grained categorization and verification.
\newblock In: Proceedings of the IEEE/CVF Conference on Computer Vision and
  Pattern Recognition (CVPR), pp. 3973--3981 (2015)

\end{thebibliography}

\end{document}